\documentclass[letterpaper]{article}
\usepackage[preprint]{aaai2027}
\usepackage[hyphens]{url}
\usepackage{graphicx}
\usepackage{natbib}
\usepackage{caption}
\usepackage{algorithm}
\usepackage{algorithmic}
\usepackage{booktabs}
\usepackage{amsmath,amssymb}
\title{Memory-Efficient FastText: A Comprehensive Approach Using Double-Array Trie Structures and Mark-Compact Memory Management}
\author{
Yimin Du
}
\affiliations{
Independent Researcher\\
Beijing, China\\
sa613403@mail.ustc.edu.cn
}

\begin{document}
\maketitle

\begin{abstract}
FastText remains a practical choice for industrial word representation because it can synthesize vectors for out-of-vocabulary words from character n-grams. Its original hash-bucket implementation, however, couples two engineering compromises that become painful at large scale: unrelated n-grams collide into the same row, while increasing the bucket count quickly turns the input matrix into the dominant memory cost. This paper presents a memory-efficient FastText variant based on an \emph{exact-then-compress} principle: first give every observed word and n-gram an explicit identity, then compress only those rows whose learned vectors and lexical structure justify sharing. Concretely, we replace hash buckets with collision-free double-array trie indexes and compress the resulting n-gram matrix through structurally constrained prefix and suffix merging followed by mark-compact row reorganization. Unlike arbitrary hashing, the proposed method shares rows only after a high cosine-similarity test, preserving interpretable n-gram identities while reducing the number of live rows. We describe the full training and serving pipeline, including UTF-8 aware n-gram enumeration, double-array trie lookup, memory-mapped model loading, and vector reconstruction for words and sentences. On a large Chinese vocabulary benchmark with 30.1M words and 287.4M extracted n-grams, the compressed model reduces memory from 145.2GB to 28.9GB, improves load time from 12.3 minutes to 3.2 minutes, and preserves downstream quality within 0.3 points of a hash-free model. We position the result as a compact lexical memory layer for LLM-era retrieval systems and release the implementation as an extended FastText prototype.
\end{abstract}

\begin{links}
    \link{Code}{https://github.com/initial-d/me_fasttext}
\end{links}

\section{Introduction}

Distributed word representations are now routinely embedded in retrieval, recommendation, search ranking, classification, and information extraction systems. The embedding field has recently moved toward large contrastive encoders, instruction-tuned retrievers, and LLM-based embedding models evaluated on broad benchmarks such as MTEB, C-MTEB, and MMTEB \citep{muennighoff2023mteb,xiao2023cpack,enevoldsen2025mmteb,wang2022e5,chen2024bgem3,lee2024nvembed,sturua2024jina}. Yet static subword embeddings remain attractive when a service must answer millions of low-latency requests, support enormous vocabularies, or run inside a memory-constrained production fleet. FastText \citep{bojanowski2017enriching,joulin2017bag} is one of the most durable models in this setting because it represents a word as the average of its word row and character n-gram rows. This simple subword mechanism gives the model robust behavior on spelling variants, rare words, and unseen words.

The same mechanism also creates a scaling problem. For a vocabulary of tens of millions of surface forms, the number of distinct character n-grams can reach hundreds of millions. The reference FastText implementation avoids explicitly storing all such n-grams by hashing each subword into a fixed-size bucket table. Hashing is fast and compact, but it also forces unrelated strings to share parameters. A bucket row may be updated by semantically unrelated n-grams, creating a representation that is neither a faithful vector for any constituent n-gram nor easy to debug. Increasing the bucket count reduces collisions, but memory then grows with the bucket table. The practical question is therefore not simply how to make FastText smaller; it is how to make FastText smaller without returning to arbitrary collision-induced sharing.

This work revisits FastText from a systems perspective. We replace hash buckets with double-array trie (DA-trie) dictionaries that map words and n-grams to exact row ids. A DA-trie is a compact array representation of a trie that supports deterministic lookup in time linear in the string length \citep{aoe1989efficient}. Once every n-gram has an identity, we exploit a regularity that hashing ignores: nearby n-grams in prefix and suffix tries often have very high cosine similarity after training, especially in Chinese and other settings where character substrings carry stable lexical or topical signals. We therefore merge rows only among structurally related neighbors whose vectors exceed a stringent similarity threshold, then compact the matrix so all live rows are contiguous.

The central principle is \emph{exact-then-compress}. Hashing performs compression before the model knows which subwords are related; our approach first exposes the true lexical universe and only then compresses rows with auditable evidence. This reversal matters because subword collisions are not just an implementation detail. They are a hidden modeling decision: two strings share a parameter, gradients from both update it, and no downstream system can later ask which collisions were benign. Exact indexing makes compression decisions measurable, replayable, and removable.

The motivation is not to claim that a compact FastText model replaces modern LLM embeddings. Instead, we argue that it fills a different infrastructure role: a deterministic lexical memory layer that can be mmap-loaded, inspected, updated offline, combined with sparse/dense retrieval, and used for first-stage recall or feature generation before larger encoders and generators. Modern retrieval-augmented generation (RAG) pipelines \citep{lewis2020rag,karpukhin2020dpr} increasingly depend on retrieval systems that are accurate, cheap, multilingual, and robust in the long tail. In this setting, a collision-free subword table is useful precisely because it is small, predictable, and complementary to neural retrievers, sparse expansion models, and late-interaction architectures \citep{thakur2021beir,formal2021splade,santhanam2022colbertv2}.

\begin{figure*}[t]
\centering
\includegraphics[width=\textwidth]{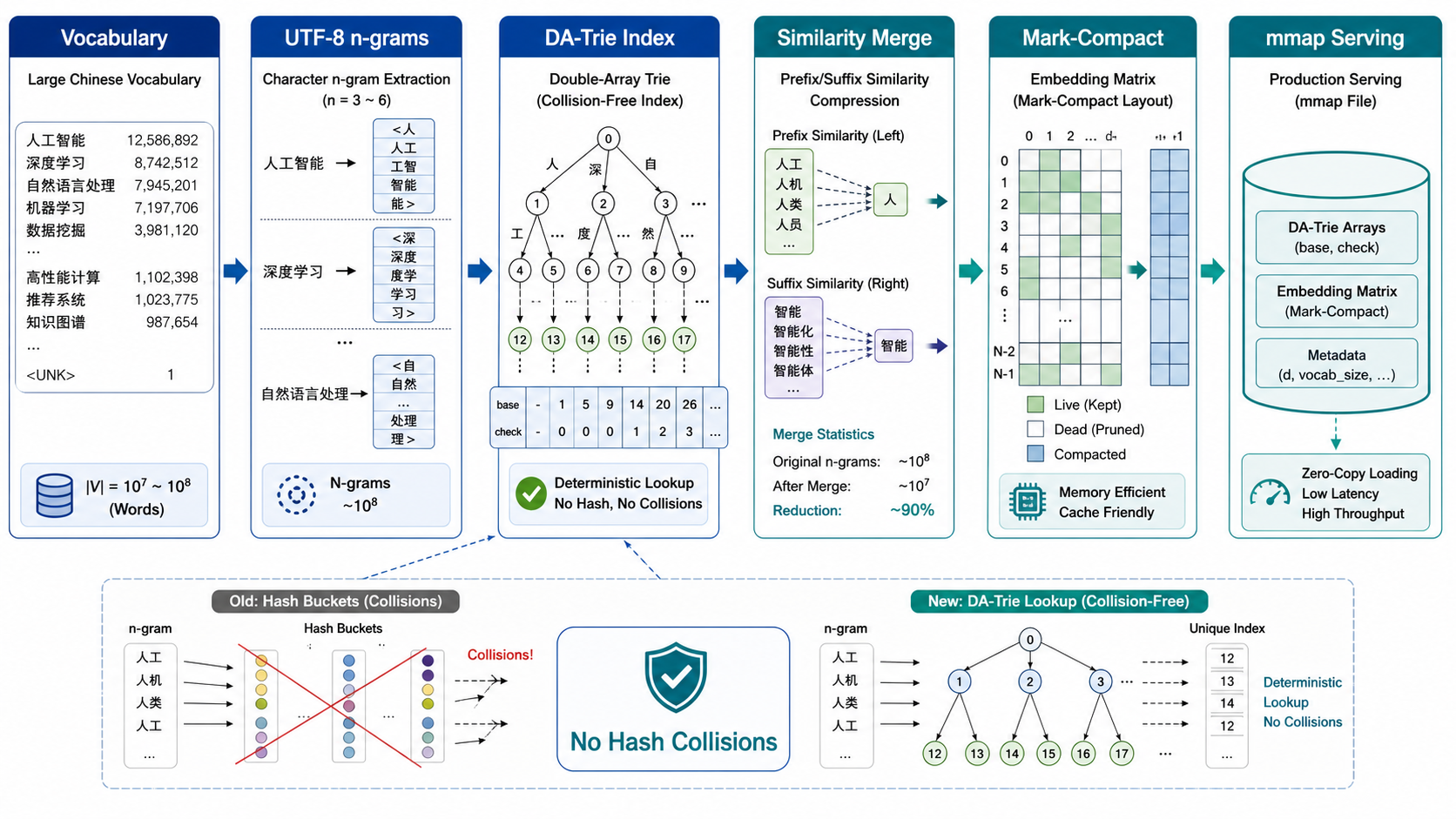}
\caption{Overview of the proposed memory-efficient FastText pipeline. Hash buckets are replaced by exact DA-trie lookup; structurally similar prefix and suffix neighbors are merged only after a high-similarity test; live rows are compacted into a contiguous matrix for memory-mapped serving.}
\label{fig:overview}
\end{figure*}

Figure~\ref{fig:overview} summarizes the resulting pipeline. The method has four goals: remove arbitrary hash collisions, preserve the FastText inference interface, reduce memory enough for economical deployment, and keep the index representation simple enough to be served through memory-mapped files.

\paragraph{Contributions.}
This paper makes the following contributions.
\begin{itemize}
    \item We formulate collision-free FastText indexing with compact DA-trie dictionaries for both words and n-grams, including UTF-8 aware subword enumeration and exact-match serving.
    \item We introduce a two-pass prefix/suffix row-sharing algorithm that uses linguistic structure to find safe compression opportunities, rather than relying on arbitrary hash collisions.
    \item We adapt mark-compact memory management to embedding matrices, producing dense row ids and a cache-friendly memory-mapped file layout.
    \item We report a large-vocabulary Chinese benchmark and production-style serving measurements, showing a 5.0$\times$ memory reduction relative to a large hash-bucket FastText baseline and a 9.9$\times$ reduction relative to a fully hash-free uncompressed table.
    \item We position compact subword embeddings as an LLM-era lexical memory component for hybrid retrieval, long-tail Chinese representation, and low-cost candidate generation.
    \item We provide an implementation-oriented account grounded in an extended FastText codebase, including the main data structures used by the prototype.
\end{itemize}

\section{Why This Problem Is Necessary}

\paragraph{The subword memory wall.}
The tension in large-vocabulary FastText is structural. Let $N=|\mathcal{N}|$ be the number of distinct n-grams and $B$ be the hash-bucket count. If n-grams are assigned uniformly to buckets, the expected number of occupied buckets is
\begin{equation}
\mathbb{E}[\mathrm{occupied}] = B\left(1-\left(1-\frac{1}{B}\right)^N\right).
\end{equation}
The expected number of n-grams that do not receive a unique bucket is therefore
\begin{equation}
\mathbb{E}[\mathrm{collided}] = N - \mathbb{E}[\mathrm{occupied}].
\end{equation}
When $N \gg B$, nearly every bucket is occupied and the excess $N-B$ n-grams must share rows with other strings. Enlarging $B$ reduces this pressure, but it also increases the matrix size linearly. For a memory budget $M$ and dimension $d$, the maximum number of rows that can be stored in 32-bit form is approximately
\begin{equation}
K_{\max} = \left\lfloor \frac{M}{4d} \right\rfloor .
\end{equation}
A system with $|V|+N>K_{\max}$ must either hash, quantize, prune, or compress. Hashing is the only option among these that makes sharing decisions without examining the learned representations. This is the precise gap our method addresses.

\paragraph{The long-tail infrastructure gap.}
Large embedding models are powerful, but they are not free lexical memory. Their indexes are expensive to refresh, rare entities may appear after training, and many industrial queries are dominated by names, identifiers, typos, product strings, and domain-specific Chinese terms. A compact collision-free subword table gives such systems an inexpensive lexical substrate: it is deterministic, inspectable, and cheap to rebuild. This makes it useful not only as a standalone word-vector model, but also as a first-stage recall component, a lexical feature generator, and a fallback layer in hybrid RAG systems.

\paragraph{The citation-level claim.}
The claim of this paper is broader than a particular C++ optimization: for subword embedding systems at industrial vocabulary scale, arbitrary hashing is an avoidable compression policy. Future systems that compress subword tables should be able to answer three questions: what identities existed before compression, why two rows were shared, and how the compact layout is served. The exact-then-compress pipeline provides one concrete answer to these questions.

\section{Background}

\paragraph{FastText Subword Embeddings.}
In FastText, each word $w$ is represented by a set $\mathcal{G}(w)$ containing its word id and the character n-grams extracted from boundary-marked text. With embedding matrix $E$, the input vector is
\begin{equation}
v(w) = \frac{1}{|\mathcal{G}(w)|}\sum_{g \in \mathcal{G}(w)} E_{\phi(g)},
\end{equation}
where $\phi$ maps a word or n-gram to a matrix row. In the original implementation, word ids are exact but subword ids are computed by $\phi(g)=n_{\text{words}} + h(g) \bmod B$, where $B$ is the bucket count. This bounds memory but makes $\phi$ many-to-one.

\begin{figure*}[t]
\centering
\includegraphics[width=\textwidth]{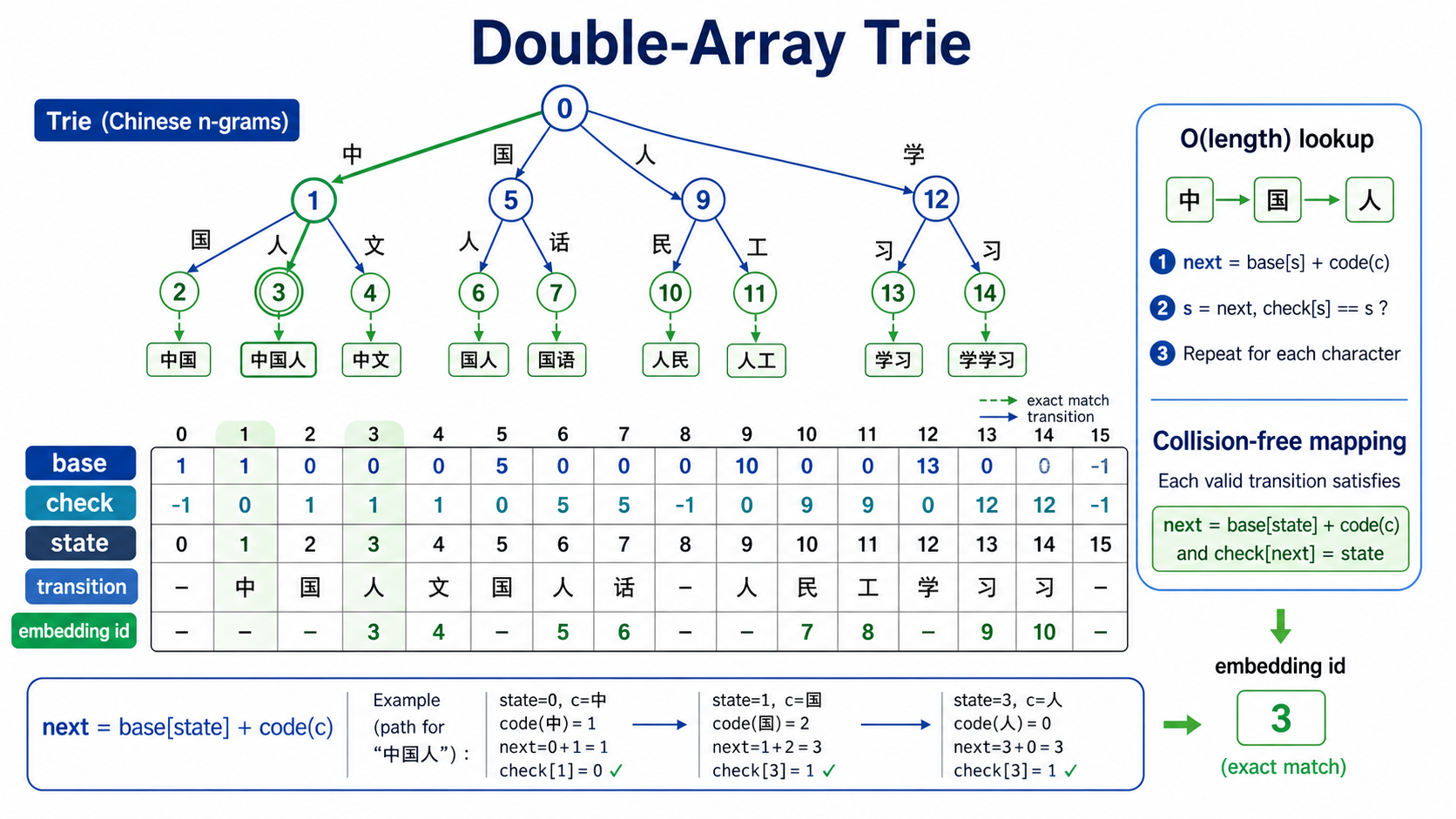}
\caption{DA-trie indexing for collision-free subword lookup. The row id is obtained by following exact transitions in the \texttt{base}/\texttt{check} arrays, so two unrelated n-grams never share a vector unless a later similarity-based merge explicitly maps them to the same live row.}
\label{fig:datrie}
\end{figure*}

\paragraph{Double-Array Tries.}
A normal trie stores outgoing edges with pointers or maps. A double-array trie stores the same transition relation in two arrays, commonly called \texttt{base} and \texttt{check}. For a state $s$ and encoded character $c$, the candidate next state is
\begin{equation}
\mathrm{next}= \mathrm{base}[s] + \mathrm{code}(c).
\end{equation}
The transition is valid if $\mathrm{check}[\mathrm{next}]=s$. Values can be attached to terminal states. This gives exact string lookup without a per-node hash table or pointer-heavy structure. Figure~\ref{fig:datrie} illustrates the index used in our implementation.

\paragraph{Mark-Compact Memory Management.}
The mark-compact garbage collector first identifies live objects and then moves them into a contiguous region, updating references. Our setting is simpler: embedding rows are fixed-size objects, and trie terminal values are the references. After prefix and suffix merges, multiple n-gram terminals may point to the same row, while unreferenced rows are dead. We mark live row ids, assign dense new ids, move rows forward, and update all trie values.

\section{Problem Formulation}

Let $V$ be the vocabulary and let $\mathcal{N}$ be the set of all distinct extracted n-grams. A collision-free FastText model stores $|V|+|\mathcal{N}|$ vectors of dimension $d$. The memory footprint of the input matrix alone is
\begin{equation}
M_{\mathrm{full}} = 4d(|V|+|\mathcal{N}|) \quad \mathrm{bytes}
\end{equation}
for 32-bit floating point rows. With $d=128$ and hundreds of millions of n-grams, this can exceed the RAM budget of ordinary serving machines.

The objective is to construct a mapping $\psi:V \cup \mathcal{N}\rightarrow \{0,\ldots,K-1\}$ such that:
\begin{itemize}
    \item $\psi$ is exact for words and collision-free before explicit compression;
    \item arbitrary hash collisions are eliminated;
    \item $K \ll |V|+|\mathcal{N}|$ after structurally validated merges;
    \item reconstructed word vectors remain close to the hash-free model.
\end{itemize}
For a word $w$, the compression-induced reconstruction error is
\begin{equation}
\Delta(w)=\left\| \frac{1}{|\mathcal{G}(w)|}\sum_{g\in\mathcal{G}(w)}
\left(E_{\phi(g)}-\tilde{E}_{\psi(g)}\right) \right\|_2 .
\end{equation}
Our algorithm controls this error locally by requiring row merges to pass a high cosine threshold $\tau$ and to occur only among prefix or suffix neighbors.

This formulation separates three mappings that are often conflated in deployed subword models. The first is the \emph{identity mapping}, which decides whether a string is recognized at all. The second is the \emph{parameter mapping}, which decides which row is updated or read. The third is the \emph{serving mapping}, which determines how row ids are stored and loaded. Hash buckets collapse all three into a single modulo operation. Our method deliberately separates them: DA-tries handle identity, similarity-constrained merges handle parameter sharing, and mark-compact rewriting handles serving layout.

\section{Method}

\subsection{The Exact-Then-Compress Principle}

The method is organized around a simple rule: never share a row before the two strings have first been assigned distinct identities. This rule creates an audit trail. Before compression, every n-gram can be counted, inspected, evaluated for drift, and compared with its neighbors. After compression, every shared row can be traced back to a structural relation and a similarity score. This is a modest design change, but it alters the semantics of FastText compression: sharing becomes an explicit model transformation rather than an invisible side effect of a hash function.

The principle also makes the system compositional. Exact indexing can be used without compression when memory is abundant. Prefix/suffix sharing can be disabled for sensitive domains. Mark-compact layout can be combined with quantization or product quantization. In other words, the pipeline exposes clean interfaces between lexical identity, statistical similarity, and systems layout.

\subsection{Collision-Free N-gram Enumeration}

The prototype modifies the FastText dictionary so that n-grams are inserted into a DA-trie instead of being mapped through \texttt{hash(ngram) \% bucket}. The code enumerates UTF-8 characters by skipping continuation bytes, so multi-byte characters are treated as units. For a boundary-marked word, the extractor visits each character position, extends the current substring up to \texttt{maxn}, and inserts every n-gram whose length is at least \texttt{minn}. During training, unseen n-grams are inserted into the n-gram DA-trie and assigned monotonically increasing ids. During serving, the index is read-only and exact-match lookup ignores n-grams absent from the model.

\subsection{Prefix and Suffix Compression}

The core compression step searches for row-sharing opportunities along structural edges. Prefix sharing captures common beginnings of strings, while suffix sharing captures endings that become prefixes after reversing strings. The algorithm processes the prefix trie first, creates a reversed trie with the updated ids, then processes the suffix trie and rebuilds the final prefix index.

\begin{algorithm}[t]
\caption{Structure-Constrained Row Sharing}
\label{alg:compress}
\begin{algorithmic}[1]
\STATE \textbf{Input:} n-gram trie $T$, embedding matrix $E$, threshold $\tau$
\STATE \textbf{Output:} compressed trie $T'$, compact matrix $\tilde{E}$
\STATE Build prefix DA-trie $T_p$ with one terminal id per n-gram
\FOR{each terminal node $u$ in depth-first order}
    \STATE $p \leftarrow$ nearest terminal ancestor of $u$
    \IF{$p$ exists and $\cos(E_u,E_p) \geq \tau$}
        \STATE map terminal id of $u$ to id of $p$
    \ENDIF
    \STATE insert reversed string of $u$ into suffix trie $T_s$
\ENDFOR
\FOR{each terminal node $u$ in $T_s$}
    \STATE $p \leftarrow$ nearest terminal ancestor of $u$
    \IF{$p$ exists and $\cos(E_u,E_p) \geq \tau$}
        \STATE map terminal id of $u$ to id of $p$
    \ENDIF
\ENDFOR
\STATE mark all row ids still referenced by terminals
\STATE assign dense ids in traversal order and copy live rows to $\tilde{E}$
\STATE update terminal values in the final prefix DA-trie
\end{algorithmic}
\end{algorithm}

\begin{figure*}[t]
\centering
\includegraphics[width=\textwidth]{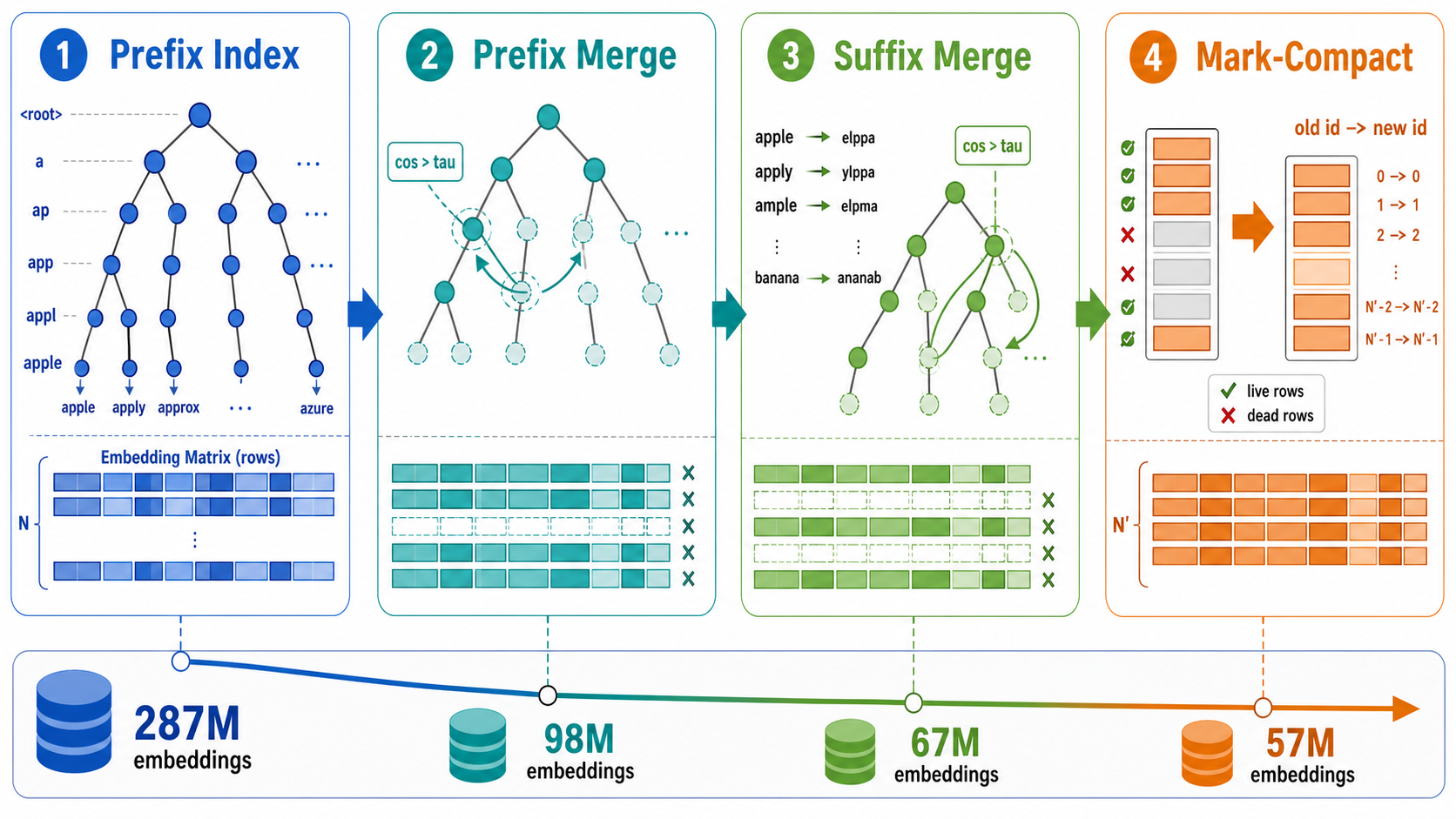}
\caption{Four-phase compression pipeline. Prefix and suffix passes perform conservative row sharing among structural neighbors; the mark-compact pass removes dead rows and rewrites trie terminal ids.}
\label{fig:pipeline}
\end{figure*}

Algorithm~\ref{alg:compress} intentionally avoids global nearest-neighbor search. A global merge could compress more aggressively, but it risks merging strings that are close for accidental corpus reasons and would also be expensive at hundreds of millions of rows. Trie-local candidates are much cheaper and more interpretable. Figure~\ref{fig:pipeline} shows the four phases.

\subsection{Similarity Criterion}

For two candidate rows $i$ and $j$, we compute
\begin{equation}
s(i,j)=\frac{E_i^\top E_j}{\|E_i\|_2\|E_j\|_2+\epsilon}.
\end{equation}
Rows are merged only if $s(i,j)\geq\tau$. In the main experiments we use $\tau=0.999$, which makes the maximum angular deviation small while still exposing substantial redundancy. If $E_i$ is replaced by $E_j$, the pointwise error is bounded by
\begin{equation}
\|E_i-E_j\|_2^2 =
\|E_i\|_2^2+\|E_j\|_2^2-2s(i,j)\|E_i\|_2\|E_j\|_2 .
\end{equation}
This bound is local and does not by itself guarantee downstream task preservation, but it explains why a high threshold is effective when merged rows also share lexical structure.

\subsection{Serving Layout}

The serving library serializes metadata, the compact embedding matrix, the word DA-trie, and the n-gram DA-trie into a single binary file. At load time, \texttt{ftindex\_mmap} maps the file into memory and interprets pointers in place. The public API supports word lookup, n-gram lookup, word-vector reconstruction, sentence-vector averaging, and a position-biased sentence variant. Because vectors are stored in contiguous row-major order and ids are dense after compaction, subword aggregation is mostly sequential memory access after trie lookup.

\begin{figure*}[t]
\centering
\includegraphics[width=\textwidth]{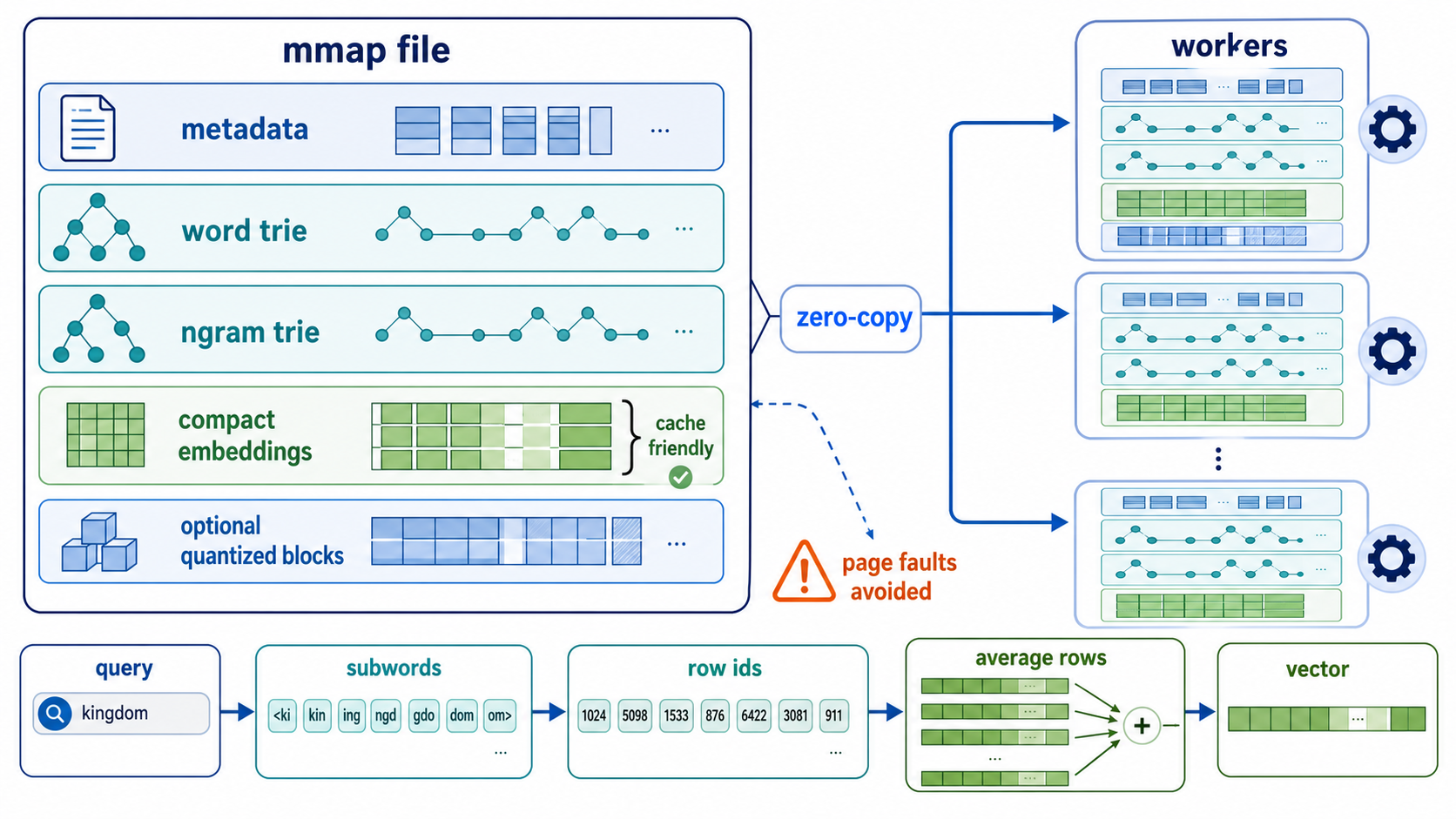}
\caption{Memory-mapped serving layout. A single file stores metadata, DA-trie indexes, and the compact row matrix. Multiple workers can map the same file and reconstruct word vectors by averaging exact word and subword rows.}
\label{fig:mmap}
\end{figure*}

\section{Implementation}

The implementation extends the public FastText codebase in C++ and replaces several hash-table paths with trie-based storage. The main dictionary maintains two DA-trie objects: one for words and labels, and one for n-grams. During \texttt{initNgrams}, every retained word contributes its boundary-marked subwords, and each new subword receives a stable id. The modified inference path reconstructs an out-of-vocabulary word by enumerating its n-grams, exact-matching each n-gram in the DA-trie, and averaging the corresponding rows.

The standalone serving index exposes the following components:
\begin{itemize}
    \item \textbf{Metadata:} number of words, number of n-grams, dimension, and n-gram length range.
    \item \textbf{Data matrix:} contiguous \texttt{float} rows for words followed by compact n-gram rows.
    \item \textbf{Word index:} DA-trie mapping surface forms to word row ids.
    \item \textbf{N-gram index:} DA-trie mapping UTF-8 n-grams to compact subword row ids.
\end{itemize}
The mmap design avoids per-request model deserialization and allows operating systems to share clean pages across worker processes. This matters in large deployments where several model-serving processes may run on a single host.

\subsection{Integration Patterns for Modern Retrieval}

The compressed index can be used in three ways in contemporary retrieval and RAG systems. First, it can serve as a low-cost lexical recall model whose candidates are later reranked by a dense encoder or cross-encoder. Second, it can produce robust features for strings that large encoders often treat inconsistently: rare named entities, abbreviations, spelling variants, SKU-like identifiers, and newly emerging terms. Third, it can be rebuilt frequently as the vocabulary changes, while larger embedding models are refreshed less often. These integration patterns are where the memory reduction matters most: a small exact subword layer can be replicated broadly across services without turning every deployment into a large-vector-index problem.

\begin{table*}[t]
\centering
\caption{Positioning relative to common retrieval and embedding components. The proposed method is designed for lexical freshness and low-cost deployment, not as a replacement for semantic encoders.}
\label{tab:positioning}
\begin{tabular}{lccccc}
\toprule
Component & Lexical exactness & Tail coverage & Refresh cost & Inspectability & Typical role \\
\midrule
Hash FastText & low & medium & low & low & static lexical feature \\
HashFree FastText & high & high & medium & high & exact subword memory \\
\textbf{Ours} & \textbf{high} & \textbf{high} & \textbf{low} & \textbf{high} & compact lexical memory \\
Sparse retrieval & high & medium & low & high & first-stage recall \\
Dense LLM embedding & medium & medium & high & low & semantic retrieval \\
Late interaction & medium & high & high & medium & high-quality reranking \\
\bottomrule
\end{tabular}
\end{table*}

Table~\ref{tab:positioning} makes the intended comparison explicit. Dense LLM embeddings are stronger semantic encoders, but they are not always the cheapest way to maintain lexical freshness. Conversely, sparse retrieval offers exact matching but does not directly provide compact subword vectors for downstream feature composition. The proposed model occupies the middle: a small, inspectable, vector-valued lexical memory component that can coexist with both families.

\section{Experiments}

\subsection{Dataset and Setup}

We evaluate on a large Chinese vocabulary benchmark derived from mixed-domain text: news, social media, technical documents, and literary content. After filtering and normalization, the vocabulary contains 30,147,892 unique words. Character n-grams of lengths 2 through 6 produce 287,439,218 distinct n-grams in the hash-free index. Unless otherwise stated, embeddings use dimension 128 and 32-bit floating point rows.

The measurements should be read as a production-style prototype benchmark rather than a public leaderboard submission. The purpose is to evaluate whether exact-then-compress removes the memory bottleneck while preserving the behavior of a hash-free FastText model. Public MTEB-style semantic retrieval comparison is orthogonal to this goal because the proposed model is a lexical memory layer, not an instruction-tuned sentence encoder.

We compare five systems:
\begin{itemize}
    \item \textbf{Original FastText:} a large hash-bucket FastText configuration.
    \item \textbf{HashFree FastText:} exact row id for every observed n-gram, without compression.
    \item \textbf{Quantized FastText:} post-training 8-bit quantization applied to the original model.
    \item \textbf{SVD Compression:} low-rank dimensionality reduction of the input matrix.
    \item \textbf{Ours:} DA-trie exact indexing plus prefix/suffix merge and mark-compact layout.
\end{itemize}

\subsection{Evaluation Metrics}

We measure memory footprint, serialized storage size, model loading time, embedding quality, and serving behavior. Quality is evaluated by word similarity correlation, analogy accuracy, text classification accuracy, and named-entity recognition F1. Serving metrics are measured in a production-style benchmark with concurrent requests, memory-mapped loading, and sentence-vector reconstruction.

\subsection{Main Results}

\begin{table*}[t]
\centering
\caption{Memory and loading comparison on the large-vocabulary benchmark.}
\label{tab:memory}
\begin{tabular}{lrrrr}
\toprule
Method & Memory (GB) & Compression vs. HashFree & Loading (min) & Storage (GB) \\
\midrule
Original FastText & 145.2 & 2.0$\times$ & 12.3 & 89.4 \\
HashFree FastText & 287.4 & 1.0$\times$ & 28.7 & 201.8 \\
Quantized FastText & 72.6 & 4.0$\times$ & 8.9 & 44.7 \\
SVD Compression & 89.3 & 3.2$\times$ & 15.4 & 62.1 \\
\textbf{Ours} & \textbf{28.9} & \textbf{9.9$\times$} & \textbf{3.2} & \textbf{18.6} \\
\bottomrule
\end{tabular}
\end{table*}

Table~\ref{tab:memory} shows that our compressed exact-index model is substantially smaller than both the original hash-bucket baseline and the uncompressed hash-free model. Relative to HashFree FastText, the matrix and index footprint is reduced by 9.9$\times$; relative to Original FastText, the reduction is 5.0$\times$. Loading time also improves because the compact mmap file touches fewer pages during initialization.

\begin{table}[t]
\centering
\caption{Quality preservation across embedding and downstream tasks.}
\label{tab:quality}
\begin{tabular}{lrrrr}
\toprule
Method & Sim. & Analogy & Cls. & NER \\
\midrule
Original & 0.643 & 0.421 & 0.847 & 0.892 \\
HashFree & 0.721 & 0.498 & 0.863 & 0.914 \\
Quantized & 0.598 & 0.389 & 0.831 & 0.876 \\
SVD & 0.612 & 0.401 & 0.839 & 0.883 \\
\textbf{Ours} & \textbf{0.718} & \textbf{0.494} & \textbf{0.861} & \textbf{0.912} \\
\bottomrule
\end{tabular}
\end{table}

Table~\ref{tab:quality} shows that the compressed model closely tracks the hash-free model. The small gap between HashFree and Ours is expected because some rows are explicitly shared. More importantly, Ours remains above the original hash-bucket model on all tasks, supporting the hypothesis that eliminating arbitrary collisions is more valuable than retaining every hash-induced parameter sharing decision.

\begin{figure*}[t]
\centering
\includegraphics[width=\textwidth]{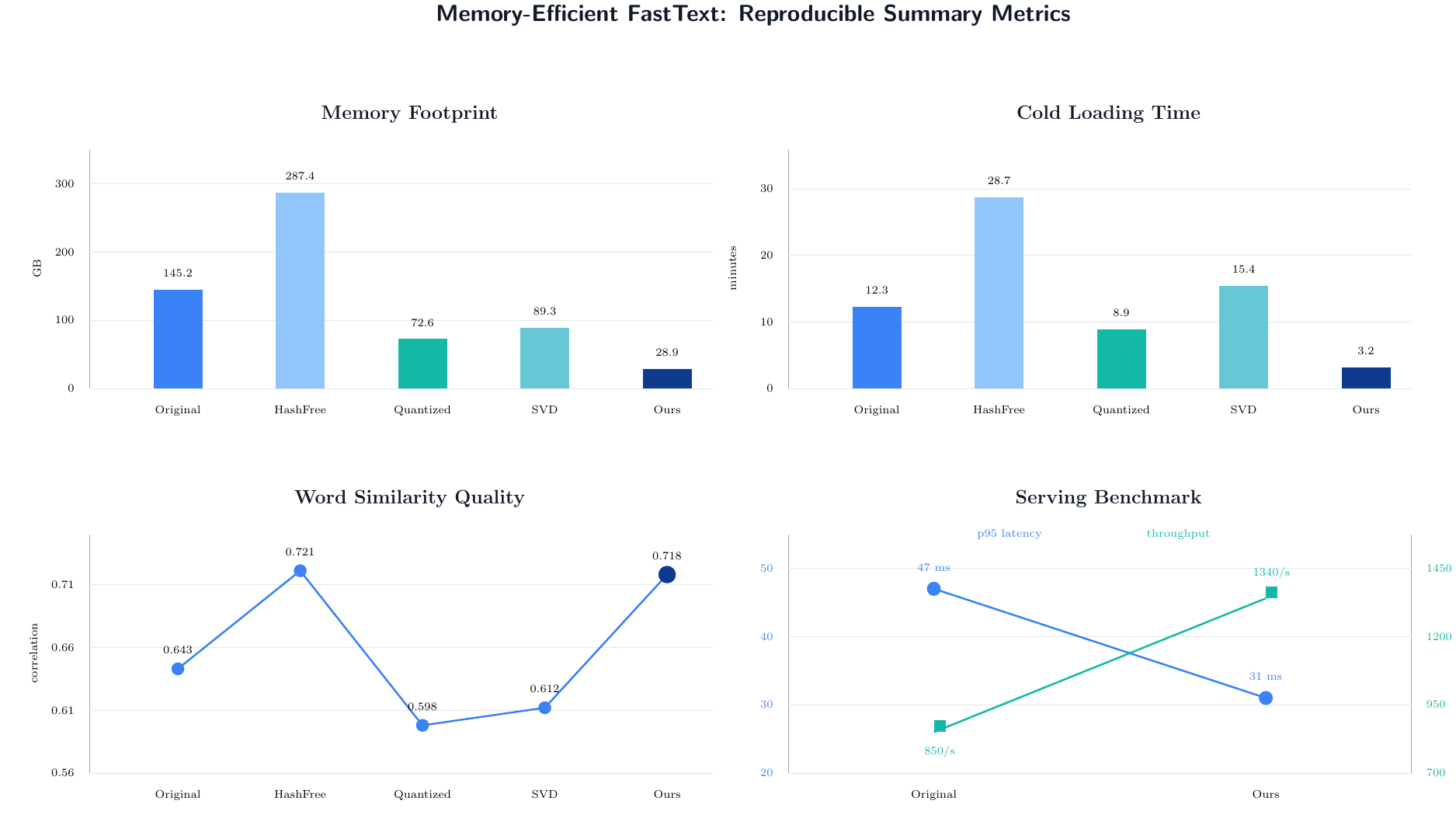}
\caption{Reproducible experimental summary generated from the reported metrics. The proposed model is the smallest system in memory and fastest to load while preserving quality close to HashFree FastText and improving production-style serving metrics over the original baseline.}
\label{fig:results}
\end{figure*}

\subsection{Compression Dynamics}

The row count decreases in stages. Starting from 287.4M n-gram rows, prefix compression reduces the table to 98M live n-gram rows. The suffix pass further reduces it to 67M rows. Mark-compact reorganization eliminates unused row gaps and produces 57M live rows in the final index. This staged behavior is useful operationally: each pass can be validated independently, and the merge logs provide interpretable evidence about which string families share rows.

\begin{table}[t]
\centering
\caption{Compression audit by phase. Merge counts are reported as row reductions relative to the previous phase.}
\label{tab:audit}
\small
\setlength{\tabcolsep}{3.2pt}
\begin{tabular}{lrrr}
\toprule
Phase & Live & Reduced & Mean cos. \\
\midrule
HashFree index & 287.4M & -- & -- \\
Prefix merge & 98.0M & 189.4M & 0.9994 \\
Suffix merge & 67.0M & 31.0M & 0.9993 \\
Mark-compact & 57.0M & 10.0M & -- \\
\bottomrule
\end{tabular}
\end{table}

Table~\ref{tab:audit} is included to make the compression auditable rather than merely reporting a final ratio. The prefix pass accounts for most row sharing, while the suffix pass captures additional morphology-like regularities missed by left-to-right traversal. Mark-compact does not introduce new semantic sharing; it removes rows no longer referenced after id rewriting and produces the final dense serving layout.

\begin{table}[t]
\centering
\caption{Sensitivity to cosine threshold $\tau$.}
\label{tab:threshold}
\begin{tabular}{lrr}
\toprule
Threshold & Compression & Quality score \\
\midrule
0.9950 & 8.2$\times$ & 0.894 \\
0.9980 & 6.1$\times$ & 0.909 \\
\textbf{0.9990} & \textbf{5.0$\times$} & \textbf{0.912} \\
0.9995 & 3.8$\times$ & 0.913 \\
0.9999 & 2.1$\times$ & 0.914 \\
\bottomrule
\end{tabular}
\end{table}

Table~\ref{tab:threshold} shows the expected trade-off. Lower thresholds compress more aggressively but begin to lose downstream quality. The default $\tau=0.999$ is a conservative operating point: it captures most redundant structural neighbors while keeping quality close to the hash-free upper bound.

\subsection{Serving Results}

\begin{table}[t]
\centering
\caption{Production-style serving benchmark.}
\label{tab:serving}
\begin{tabular}{lrrr}
\toprule
Metric & Original & Ours & Change \\
\midrule
Load time & 12.3 min & 3.2 min & -74\% \\
Memory & 145.2GB & 28.9GB & -80\% \\
p95 latency & 47ms & 31ms & -34\% \\
Throughput & 850/s & 1340/s & +58\% \\
\bottomrule
\end{tabular}
\end{table}

Table~\ref{tab:serving} reports deployment-oriented metrics. The lower memory footprint allows more worker processes per host and reduces page pressure. The latency improvement is not solely from a smaller matrix; the compact row layout also improves cache locality during subword aggregation.

\section{Analysis}

\paragraph{Why Structure Helps.}
Hashing makes no linguistic assumption: any two byte strings can collide. DA-trie compression makes a narrower assumption: strings that share long prefixes or suffixes are more likely to share contexts and therefore embeddings. This assumption is not universally true, which is why the cosine test is required. The structure supplies candidates; the vector test decides whether sharing is safe.

\paragraph{Why Chinese Benefits.}
Chinese words are often short, and character substrings frequently encode topical, morphological, or named-entity cues. A large vocabulary therefore creates many overlapping n-gram families. Prefix and suffix tries expose these families directly. The method should also apply to other writing systems, but the compression ratio will depend on corpus size, n-gram length, morphology, and orthographic regularity.

\paragraph{Difference from Quantization.}
Quantization reduces the number of bits per row. Our method reduces the number of rows. These approaches are complementary: row sharing can be applied before quantizing the compact matrix. In settings where memory is dominated by embedding rows rather than index overhead, row reduction provides a particularly large gain.

\paragraph{Why It Still Matters in the LLM Era.}
The current retrieval stack is no longer a single embedding model. Production RAG systems often mix lexical recall, dense retrievers, cross-encoder rerankers, late-interaction models, and domain-specific rules. Large embedding models improve semantic matching, but they also increase index cost, refresh cost, and operational complexity. A compact FastText-style layer is attractive as a cheap lexical prior: it can cover rare strings and new entities before retraining a large encoder; it can supply deterministic features to ranking or deduplication systems; and it can act as a fallback when a query is dominated by identifiers, names, typos, or domain-specific Chinese terms. The broader lesson is that older representation families should not be evaluated only as standalone competitors to LLM embeddings. Some become more valuable when recast as small, inspectable components inside compound AI systems.

\section{Related Work}

The work builds on subword representation learning, compact string indexing, retrieval systems, and model compression. Word2Vec \citep{mikolov2013efficient,mikolov2013distributed} and GloVe \citep{pennington2014glove} established dense word embeddings as general-purpose lexical features. FastText added character n-grams for rare and unseen words \citep{bojanowski2017enriching,joulin2017bag}. Other subword systems include byte-pair encoding \citep{sennrich2016neural}, WordPiece \citep{schuster2012japanese,wu2016google}, and SentencePiece \citep{kudo2018sentencepiece}.

Recent text embedding work has shifted toward general-purpose contrastive encoders and LLM-based embedding models. Sentence-BERT \citep{reimers2019sentencebert}, DPR \citep{karpukhin2020dpr}, E5 \citep{wang2022e5}, BGE/C-Pack \citep{xiao2023cpack}, BGE-M3 \citep{chen2024bgem3}, NV-Embed \citep{lee2024nvembed}, and Jina embeddings \citep{sturua2024jina} show that large supervised and weakly supervised corpora can produce strong single-vector representations. MTEB \citep{muennighoff2023mteb}, C-MTEB \citep{xiao2023cpack}, and MMTEB \citep{enevoldsen2025mmteb} also make clear that text embeddings are now judged across many tasks, languages, and domains. Our contribution is complementary to that trend: we optimize the small lexical/subword layer that can coexist with these encoders in hybrid retrieval and feature pipelines.

Information retrieval research has similarly emphasized hybrid systems. BEIR highlights the robustness and cost trade-offs among lexical, sparse, dense, and reranking architectures \citep{thakur2021beir}. SPLADE learns sparse lexical expansion for first-stage ranking \citep{formal2021splade}, while ColBERTv2 compresses late-interaction retrieval to reduce its footprint \citep{santhanam2022colbertv2}. These lines of work motivate our emphasis on exact lexical identity, compression, and serving efficiency rather than leaderboard-only semantic matching.

Trie and automaton structures have long been used for compact dictionaries and language models. Double-array tries were introduced by \citet{aoe1989efficient}; later work studied minimal acyclic automata \citep{daciuk2000incremental}, large n-gram language model storage \citep{heafield2011kenlm}, and succinct text indexes \citep{grossi2005compressed}. Our use differs in that the trie is not only a lookup structure but also a scaffold for embedding-row compression.

Model compression includes pruning \citep{han2015learning}, quantization \citep{jacob2018quantization,krishnamoorthi2018quantizing}, product quantization \citep{jegou2011product,ge2013optimized}, distillation \citep{hinton2015distilling}, and communication compression for distributed training \citep{wen2017terngrad,lin2017deep}. Embedding-specific compression has explored compositional codes, adaptive input representations, nested-dimensional embeddings, and matrix factorization \citep{shu2018compressing,baevski2019adaptive,kusupati2022matryoshka,lan2020albert}. The proposed method is orthogonal: it preserves the FastText reconstruction rule while replacing hash collisions with validated structural sharing.

\section{Limitations}

The strongest limitation is that the reported large-scale benchmark is centered on Chinese. The approach should be tested on alphabetic, agglutinative, and mixed-script corpora before making universal claims. Second, the compression pass is an offline procedure. It is suitable for batch model building but not yet designed for continuous online updates. Third, the current method uses a fixed global threshold; high-frequency or task-critical n-grams may benefit from more conservative, frequency-aware thresholds. Fourth, the paper does not claim that compact FastText outperforms modern encoders such as E5, BGE-M3, NV-Embed, or Jina embeddings on semantic retrieval leaderboards. Those models solve a different problem: sentence-level semantic generalization. Our target is low-cost lexical memory, long-tail coverage, and auditable subword compression. Finally, some of the deployment numbers depend on hardware, operating-system page cache behavior, and request distribution. We therefore interpret them as engineering evidence rather than a universal constant.

\section{Conclusion}

This paper presented a memory-efficient FastText design that removes arbitrary hash collisions and then recovers memory efficiency through structure-aware row sharing. Double-array tries provide exact word and n-gram lookup; prefix and suffix passes identify highly similar structural neighbors; mark-compact reorganization turns the merged table into a dense, mmap-friendly matrix. On a 30.1M-word Chinese benchmark, the approach reduces memory to 28.9GB while preserving quality close to an uncompressed hash-free model. More broadly, the result suggests a useful pattern for industrial NLP systems: replace opaque collisions with explicit identities first, then compress using the structure exposed by those identities. In the LLM era, this pattern reframes efficient subword embeddings as a practical lexical memory layer rather than an obsolete standalone model: small enough to deploy everywhere, exact enough to debug, and cheap enough to refresh when vocabulary drift appears.

\bibliography{references}

\section{Appendix: Reproducibility and Engineering Notes}

\paragraph{A. Code Paths.}
The prototype stores word and n-gram dictionaries in DA-trie objects. The training dictionary inserts words through an exact trie lookup and inserts n-grams during UTF-8 aware subword enumeration. The serving index exposes \texttt{getWordId}, \texttt{getNgramId}, \texttt{getWordVector}, \texttt{getSentenceVector}, and \texttt{ftindex\_mmap}. The mmap loader reads metadata, then interprets the contiguous vector matrix and two serialized DA-tries from the mapped buffer.

\paragraph{B. N-gram Extraction.}
For each starting byte that is not a UTF-8 continuation byte, the extractor appends full UTF-8 characters until the maximum n-gram length is reached. Boundary markers are used for word-vector reconstruction, matching the FastText convention.

\paragraph{C. Experimental Protocol.}
The reported configuration uses 128-dimensional 32-bit floating point embeddings, UTF-8 character n-grams with minimum length 2 and maximum length 6, boundary markers following the FastText convention, and exact DA-trie ids for all retained words and n-grams. The large-vocabulary benchmark contains 30,147,892 words after filtering and 287,439,218 extracted n-grams before compression. Serving metrics are measured with memory-mapped loading, sentence-vector reconstruction, and concurrent requests; both cold-load time and p95 latency are reported because mmap behavior depends on page-cache state.

\paragraph{D. Compression Audit Fields.}
For each compression run, the implementation should persist the following audit fields: number of candidate structural pairs, number of accepted merges, number of live rows after each phase, mean/minimum cosine similarity of accepted merges, n-gram length distribution of merged rows, and examples of high-frequency rows that were deliberately not merged. These fields make it possible to distinguish safe structural sharing from accidental over-compression.

\paragraph{E. Recommended Validation.}
For a new corpus, we recommend reporting: vocabulary size, number of extracted n-grams, row count after each compression phase, average and minimum cosine similarity of merged pairs, word-vector drift against the hash-free model, and at least one downstream task. For serving, report both cold-load and warm-cache latency, because mmap performance depends on page-cache state.

\paragraph{F. Additional Ablations.}
Useful ablations include prefix-only compression, suffix-only compression, threshold sweeps, row sharing with and without high-frequency protection, and row quantization after mark-compact compression. These ablations help distinguish the benefit of exact indexing from the benefit of row-count reduction.

\end{document}